\documentclass[letterpaper]{article} % DO NOT CHANGE THIS
\usepackage{aaai24}  % DO NOT CHANGE THIS
\usepackage{times}  % DO NOT CHANGE THIS
\usepackage{helvet}  % DO NOT CHANGE THIS
\usepackage{courier}  % DO NOT CHANGE THIS
\usepackage[hyphens]{url}  % DO NOT CHANGE THIS
\usepackage{graphicx} % DO NOT CHANGE THIS
\usepackage{natbib}  % DO NOT CHANGE THIS AND DO NOT ADD ANY OPTIONS TO IT
\usepackage{caption} % DO NOT CHANGE THIS AND DO NOT ADD ANY OPTIONS TO IT
\usepackage{algorithm}
\usepackage{algorithmic}
\usepackage{multirow}
\usepackage{amsmath}
\usepackage{amssymb}
\usepackage{bbding}
\usepackage{xcolor}

\usepackage{newfloat}
\usepackage{listings}
\DeclareCaptionStyle{ruled}{labelfont=normalfont,labelsep=colon,strut=off} % DO NOT CHANGE THIS
\floatstyle{ruled}
\newfloat{listing}{tb}{lst}{}
\floatname{listing}{Listing}
\title{Frequency-Adaptive Pan-Sharpening with Mixture of Experts}
\author {
    Xuanhua He\textsuperscript{\rm 1,\rm 2}\thanks{Co-first authors contributed equally. $^\dagger$ Corresponding author.},
    Keyu Yan\textsuperscript{\rm 1,\rm 2}\footnotemark[1],
    Rui Li\textsuperscript{\rm 1},
    Chengjun Xie\textsuperscript{\rm 1},
    Jie Zhang\textsuperscript{\rm 1}$^{\dagger}$,
    Man Zhou\textsuperscript{\rm 3}$^{\dagger}$
}
\affiliations {
    \textsuperscript{\rm 1}Hefei Institutes of Physical Science, Chinese Academy of Sciences
    \textsuperscript{\rm 2}University of Science and Technology of China\\
    \textsuperscript{\rm 3}Nanyang Technological University\\
    { \{hexuanhua,keyu\}@mail.ustc.edu.cn, \{lirui,cjxie,zhangjie\}@iim.ac.cn,manzhountu@gmail.com}\\
}
\usepackage{bibentry}
\begin{document}

\maketitle

\begin{abstract}
Pan-sharpening involves reconstructing missing high-frequency information in multi-spectral images with low spatial resolution, using a higher-resolution panchromatic image as guidance. Although the inborn connection with frequency domain, existing pan-sharpening research has not almost investigated the potential solution upon frequency domain. To this end, we propose a novel Frequency Adaptive Mixture of Experts (FAME) learning framework for pan-sharpening, which consists of three key components: the Adaptive Frequency Separation Prediction Module, the Sub-Frequency Learning Expert Module, and the Expert Mixture Module. In detail, the first leverages the discrete cosine transform to perform frequency separation by predicting the frequency mask. On the basis of generated mask, the second with low-frequency MOE and high-frequency MOE takes account for enabling the effective low-frequency and
high-frequency information reconstruction. Followed by, the final fusion module dynamically weights high-frequency and low-frequency MOE knowledge to adapt to remote sensing images with significant content variations.  Quantitative and qualitative experiments over multiple datasets demonstrate that our method performs the best against other state-of-the-art ones and comprises a strong generalization ability for real-world scenes. Code will be made publicly at \url{https://github.com/alexhe101/FAME-Net}.
\end{abstract}
\section{Introduction}
The demand for high-resolution multispectral (HRMS) images is increasing in various industries such as agriculture, mapping services, and environmental protection. However, direct acquisition of HRMS images using satellite sensors is often not feasible due to technology and hardware limitations. Instead, a common approach is to use two distinct sensors on satellites to capture high-resolution panchromatic (PAN) and low-resolution multispectral (LRMS) images. These images are then fused through the pan-sharpening process to generate HRMS images suitable for specific applications. 
% Given the significant demand for such images in various downstream tasks, the field of pan-sharpening has gained significant attention.
\begin{figure}[!h]
    \centering
    \includegraphics[width=\linewidth]{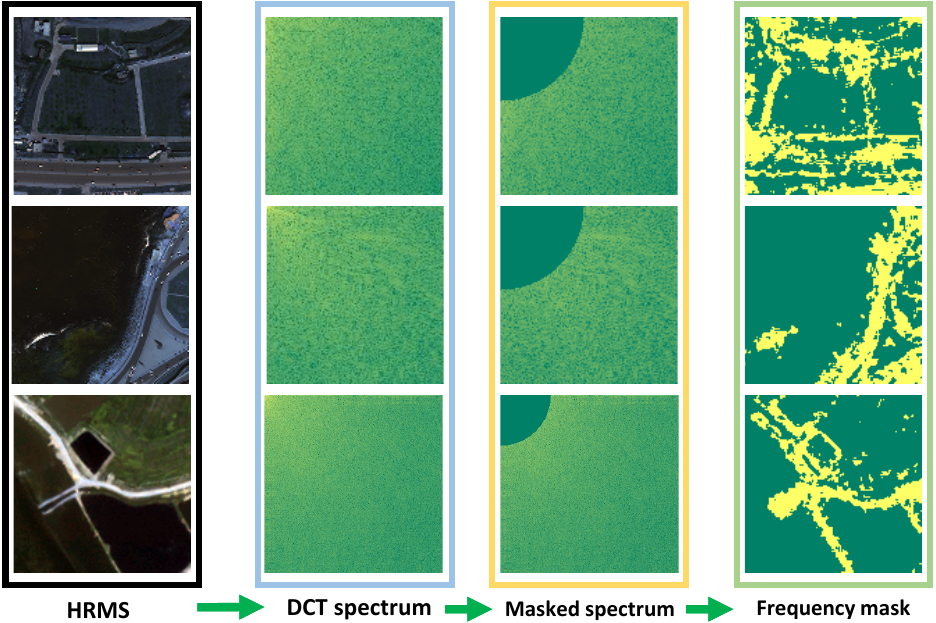}
    \caption{Generation process of frequency mask. Firstly, a discrete cosine transform is applied to the image. Then, the upper left part of the DCT spectrum is masked using manually selected thresholds. Finally, the frequency mask is generated through inverse transformation.}
    \label{fig:dctmoti}
\end{figure}

Recent years have witnessed significant progress in maintaining both spectral and spatial details over pan-sharpening as a consequence of the rapid progress of deep learning technology.
The PNN~\cite{pnn}, which takes inspiration from the SRCNN~\cite{srcnn} and employs a similar network architecture, is one of the first deep learning solutions in this field.
Despite its simplicity, the PNN has achieved remarkable improvements in various performance metrics, showcasing the strong capabilities of deep learning. 
Since then, explosive pan-sharpening networks have been proposed, leveraging advanced network architectures to attain superior visual performance.
However, existing pan-sharpening methods have overlooked the discrepancies between various frequency components of multi-spectral image and relied on a uniform approach across the entire image, limiting the potential for further spatial detail enhancement. 
As shown in the previous study~\cite{fuoli2021fourier}, there is a significant correlation between super-resolution and frequency information. Considering that pan-sharpening is essentially a super-resolution process, it is reasonable to investigate how the interaction between various frequency components in two modal images can be utilized to improve the performance of pan-sharpening models.

\textbf{Our motivation.} Our goal is to improve the performance of pan-sharpening methods by effectively recovering high-frequency information, benefiting for generating clear images with fine textures. Previous convolution network-based approaches have struggled to learn high-frequency details, as CNNs are inherently inclined towards low-frequency information~\cite{magid2021dynamic}. Recovering high-frequency information is of great importance in generating clear images.
The discrete cosine transform (DCT)~\cite{ahmed1974discrete,xie2021learning} provides a powerful tool for frequency domain analysis of images, as illustrated in Figure~\ref{fig:dctmoti}. Initially, we apply the DCT to the image to obtain the second column, where the low-frequency components are concentrated in the upper left corner of the DCT image. Subsequently, we obtain the frequency mask of the image by masking the upper left corner and employing the inverse discrete cosine transform. As shown in the fourth column of the Figure~\ref{fig:dctmoti}, the frequency mask decomposes the original image into high-frequency and low-frequency parts. This characteristic enables different modules of the network to focus on the high and low frequency parts of the image separately, explicitly encouraging the network to learn high-frequency information and generate pan-sharpened images with clear textures. Considering the significant variability in content among different remote sensing images, utilizing a dynamic network structure can enhance the model's generalization performance. The Mixture of Experts (MOE)~\cite{jordan1994hierarchical} has demonstrated efficacy in various vision tasks by leveraging expert knowledge of different parts and employing a dynamic network structure. By utilizing frequency experts to facilitate the separate learning of high- and low-frequency information and adapting to different inputs through a dynamic network structure, we can significantly enhance the performance of the pan-sharpening model.

Taking into account the above-discussed insights, we present an innovative Frequency Adaptive Mixture of Experts network, named FAMEnet. By blending the MOE technique with frequency domain information, it is able to guide the network to learn image features at different frequencies, particularly high-frequency information. Furthermore, by utilizing dynamic network structures, our proposed FAMEnet can adapt to remote sensing images with significant content variance, thereby enhancing its generalization ability.
The FAMEnet comprises three key modules: Frequency Mask predictor, Sub-frequency learning experts module, and  Experts Mixture module. The Mask predictor is responsible for generating frequency masks that segregate the image into high-frequency and low-frequency parts, thus enabling the effective processing of the image content. The Frequency experts consist of two MOE components, namely low-frequency MOE and high-frequency MOE, which are exclusively utilized for processing low-frequency and high-frequency information of the image. With the aid of the expert network, it can distinctly focus on the high- and low-frequency components of the image to achieve targeted processing.
The final experts mixture part is responsible for dynamically fusing high- and low-frequency features, as well as PAN and MS features, to adapt to remote sensing images with significant content variations. The final output is obtained by dynamically adding multiple different frequency experts. By encouraging the network to process high- and low-frequency information separately and dynamically fuse features, the generated images have clearer textures and better generalization.

Our contribution can be summarized as follows:
\begin{itemize}
    \item In this work, we devised a method that combines MOE (Mixture of Experts) with frequency domain information. In this way, we enable the network to learn and adapt to the high-frequency information present in remote sensing images in a dynamic manner.
    \item The proposed method comprises of a frequency separation mask predictor, MOE-based frequency adapatively learning module, and experts mixture module. This design allows the pan-sharpening network to effectively capture high-frequency information, leading to high-quality pan-sharpening results.
    \item Our proposed Mixture of Experts framework surpasses existing methods and achieves state-of-the-art results in pan-sharpening. The output is characterized by clear textures, accurate spectra, and strong generalization ability, as evidenced by qualitative and quantitative experiments conducted on multiple datasets.
\end{itemize}

\begin{figure*}[!ht]
    \centering
    \includegraphics[width=\textwidth]{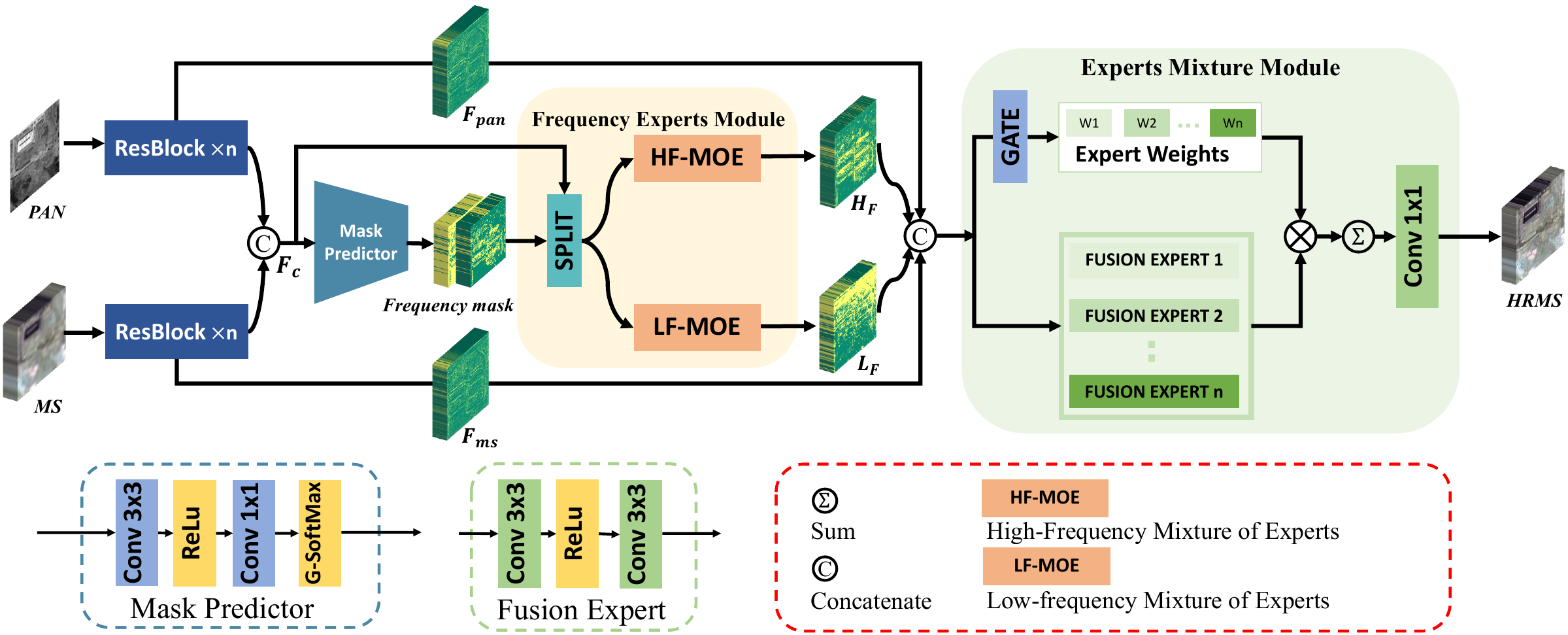}
    \caption{The overall structure of FAMEnet, which is composed of three main components: Mask predictor, Frequency Experts Module, and Experts Mixture Module.}
    \label{fig:mainfig}
\end{figure*}
\section{Related Work}
\subsection{Pan-sharpening}
A plethora of research has emerged in the community of pan-sharpening. Existing methods can be classified into traditional and deep learning-based approaches. Traditional methods include component substitution-based~\cite{IHS,Brovey,GS,GFPCA}, multi-resolution analysis-based~\cite{DWT1989,ATWT1999,LPTl,HPF}, and model-based methods~\cite{fasbender2008bayesian,tv}. However, these methods are limited by insufficient feature representation, and it is difficult to achieve satisfactory results. The success of convolutional neural networks has sparked interest in the field of pan-sharpening. PNN~\cite{pnn} was the first to introduce CNNs, which achieved significant improvements compared to traditional methods. 
 % with a lightweight network
PANNET~\cite{yang2017pannet} further improved the performance by introducing residual design. 
 % and high-frequency filtering
Since then, more complex designs and deeper networks have been used to enhance the performance of pan-sharpening task, such as MSDCNN~\cite{msdcnn} for capturing multi-scale information and SRPPNN~\cite{srppnn} with a very deep super-resolution architecture. Recently, GPPNN~\cite{gppnn} and MMNet~\cite{yan2022memory} were designed to enhance interpretability through deep unrolling methods. ARFNet~\cite{yan2022panchromatic} further explored the convergence of the unrolling process. MutNet~\cite{zhou2022mutual} introduced information theory to minimize mutual information redundancy. Inspired by the wide-spread application of transformer, INN-former~\cite{zhou2022pan} combines CNN and Transformer to promote the combination of local and global information. SFINet~\cite{zhou2022spatial} utilizes the Fourier transform to implicitly learn high-frequency features, yet it lacks explicit incentives for the network to effectively harness this information, leading to suboptimal outcomes.
These methods are limited in their ability to leverage high-frequency information, resulting in less clear generated textures.
\subsection{Mixture of experts}
MOE~\cite{jordan1994hierarchical,gross2017hard} is a widely-used technique that follows the divide-and-conquer strategy to decompose tasks into multiple parts and utilizes task-specific experts to handle them, with the final output obtained by weighting the experts. MOE's gate network can dynamically adjust the network structure according to the inputs, making it highly generalizable and widely applicable in various domains, including natural language processing~\cite{shazeer2017}, image classification~\cite{zhang2019learning}, and Re-ID~\cite{dai2021generalizable} and image fusion~\cite{cao2023multi}.
In contrast to these approaches, we propose dividing images into frequency specialists to encourage the network to capture high-frequency information. This represents the first attempt to employ MOE in the pan-sharpening community.

\section{Method}
Our proposed methodology involves utilizing DCT to generate frequency masks, and employing the FAMEnet network architecture. This section presents a brief overview of DCT, followed by a detailed description of the network structure used in this paper.
\subsection{Discrete Cosine Transform}
The DCT is a valuable tool for analyzing frequency domains, offering several advantages over the commonly used Fourier transform. With its simpler form and superior energy compression characteristics, the DCT enables the concentration of most of the energy in a few small coefficients. As a result, the DCT is highly suitable for both image compression and image frequency division.
Given an image $x \in\mathbb{R}^{H\times W}$, its cosine transformation process can be defined as:
\begin{scriptsize}
    \begin{align}
D(u,v) = \sum_{h=0}^{\rm H-1} \sum_{w=0}^{\rm W-1} x_{h,w} \cos{\left(\frac{\pi u}{H}\left(h+\frac{1}{2}\right)\right)}\cos{\left(\frac{\pi v}{W}\left(w+\frac{1}{2}\right)\right)}
\end{align}
\end{scriptsize}

As illustrated in Figure~\ref{fig:dctmoti}, the image is transformed to the frequency domain via cosine transformation. This transformation results in the majority of energy being concentrated in the upper left corner of the frequency domain, which represents the low-frequency component, with the remaining high-frequency portion elsewhere. To generate the corresponding frequency mask, we remove the low-frequency component using a manually selected radius, and then perform an inverse transformation. By utilizing frequency masks, the network can effectively concentrate on the high-frequency information and learn fine-grained details. However, since these masks are generated based on manually selected threshold values, they are not robust to different image content and are sensitive to noise. To address this issue, we propose to learn frequency masks from the network.
\subsection{Network Framework}
The overall architecture of the network is depicted in Figure~\ref{fig:mainfig}. 
 The input comprises upsampled LRMS and PAN images, from which we extract the features using ResBlock to obtain $\rm \mathbf{F}_{ms}$ and $\rm \mathbf{F}_{pan}$. These features are concatenated to yield $F_c$, which is passed through the mask predictor for frequency mask prediction. Subsequently, the frequency mask $\rm \mathbf{M} \in \mathbb{R}^{H\times W\times 2}$ and $\rm \mathbf{F}_c$ are fed into the Frequency experts module, where $\rm \mathbf{F}_c$ is separated based on the frequency mask using LF-MOE and HF-MOE to process low and high frequency features, respectively. Finally, the output of the Frequency experts module is combined with $\rm \mathbf{F}_{ms}$ and $\rm \mathbf{F}_{pan}$ and works with the frequency experts mixture to generate HRMS images.
 \begin{figure}
    \centering
    \includegraphics[width=\linewidth]{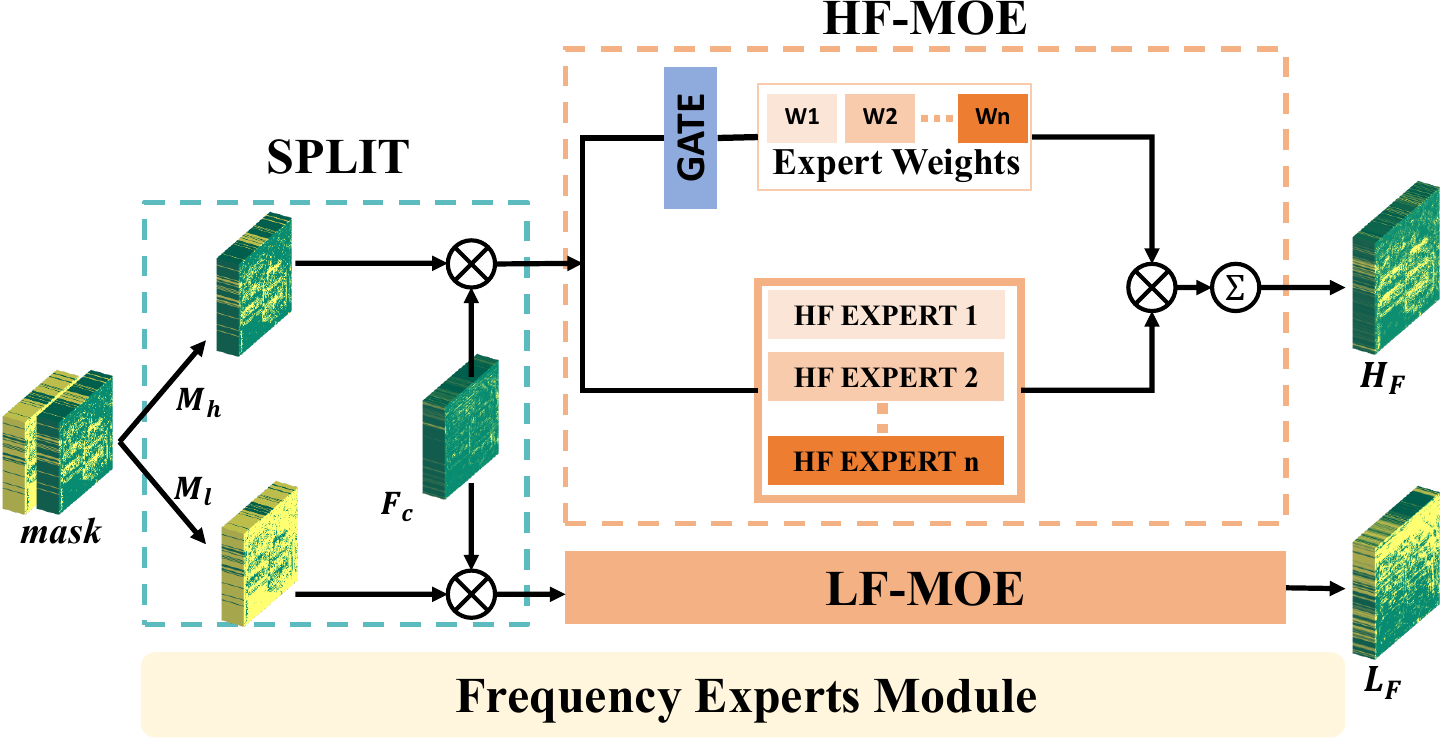}
    \caption{The architecture of the Frequency Experts Module. The frequency mask splits $F_c$ into high-frequency and low-frequency parts, which are processed separately by HF-MOE and LF-MOE.
    }
    \label{fig:fresub}
\end{figure}
\subsection{Key components}
\noindent\textbf{Mask Predictor.} Our network comprises a lightweight mask predictor module, as illustrated in Figure~\ref{fig:mainfig}. The mask predictor learns high-frequency and low-frequency components adaptively based on the image content for generating frequency masks. We utilized Gumbel-Softmax~\cite{jang2017categorical} to ensure differentiability in mask prediction.
The frequency mask $M$ for the input $\rm \mathbf{F}_c$ can be generated using the following process:
\begin{align}
    &\rm \mathbf{P} = \rm \mathbf{C}_1 \circ \mathbf{ReLu} \circ \mathbf{C}_3 (\rm \mathbf{F}_c)\\
&\rm \mathbf{M} = \rm \mathbf{GumbelSoft}(\mathbf{P})
\end{align}
Here, $\mathbf{C}_1$ and $\mathbf{C}_3$ are convolution blocks with kernel sizes of $1 \times 1$ and $3 \times 3$, respectively. The Gumbel-Softmax function $\rm \mathbf{GumbelSoft}(.)$ is applied to generate masks for the high-frequency and low-frequency components, which are represented by the 2 channels of $\rm \mathbf{M} \in \mathbb{R}^{\rm H\times W\times 2}$, and $\rm \mathbf{P} \in \mathbb{R}^{\rm H\times W\times 2}$ is intermediate feature for generating $M$. The use of the Gumbel Softmax technique ensures the differentiability of the mask generation process, as opposed to the non-differentiable $\mathbf{argmax}$ operation.

Specifically, Gumbel Softmax can be expressed as follows:
\begin{align}
 &\rm \mathbf{Z}_i =\rm  \frac{\exp((\mathbf{P}_i+\mathbf{g}_i)/{\tau})}{\sum_{c=1}^{C}\exp((\mathbf{P}_{i,c}+\mathbf{g}_{i,c})/{\tau})}\\
% &Z_i = \frac{\exp\left(\frac{P_i+g_i}{\tau}\right)}{\sum_{c=1}^{C}\exp\left(\frac{P_{i,c}+g_{i,c}}{\tau}\right)}\\
&\rm \mathbf{M}_i = \rm  \mathop{\zeta} \circ \mathop{\arg\max}\limits_{c} \mathbf{Z}_{i,c}
\end{align}
Here, $\mathbf{C}$ corresponds to the number of channels in $\mathbf{P}$, which is 2. $\mathbf{g}_i$ is the noise sampled from the Gumbel distribution, and $\tau$ is the temperature coefficient. Additionally, the function $\mathop{\zeta(.)}$ is utilized to represent the onehot encoding. Two channels are generated in $\mathbf{M}$, one for the high-frequency mask and the other for the low-frequency mask. During the backward process, we utilize the gradient of $\mathbf{Z}$ as the gradient of $\mathbf{M}$ approximately, since $\mathbf{M}$ is non-differentiable.
\noindent\textbf{Frequency Experts Module.} The targeted processing of high and low frequency components of the image can enhance the network's ability to capture frequency domain information, as illustrated in Figure~\ref{fig:fresub}. Our frequency experts module, which comprises split, LF-MOE (low frequency - mixture of experts), and HF-MOE (high frequency - mixture of experts), performs this task. The split operation separates the input into high-frequency and low-frequency components based on the frequency mask, which are then processed by HF-MOE and LF-MOE, respectively, to extract high and low-frequency features. The HF expert in HF-MOE consists of HIN (Half-Instance Normalization)~\cite{Chen_2021_CVPR} blocks, while the LF expert in LF-MOE uses 3x3 convolutions. To handle the complexity of high-frequency feature extraction, more sophisticated modules are used in the HF expert. Adaptive adjustments are made to the dynamic weights of HF-MOE and LF-MOE to choose the most suitable experts for processing the high and low frequency information that varies significantly across remote sensing images.
To define the split process, we begin with the input $\mathbf{M}$ and $\rm \mathbf{F}_c$. The process is as follows:
\begin{align}
% &\mathbf{M_h},\mathbf{M_l} = \mathop{SP}(\mathbf{M})\\
&\rm \mathbf{M}_h, \mathbf{M}_l =\rm \mathbf{M}(:,:,0),\mathbf{M}(:,:,1)\\
&\rm \mathbf{F}_h, \mathbf{F}_l = \rm \mathbf{M}_h \odot \mathbf{F}_c,  \mathbf{M}_l \odot \mathbf{F}_c
\end{align}
The mask $\mathbf{M}$ comprises two channels that correspond to high-frequency and low-frequency masks, respectively. By multiplying $\rm \mathbf{F}_c$ with these masks, we obtain the high-frequency and low-frequency components of $\rm \mathbf{F}_c$, denoted by $\rm \mathbf{F}_h$ and $\rm \mathbf{F}_l$. We then feed these two sets of features into HF-MOE and LF-MOE, respectively, to obtain high-frequency and low-frequency features $\rm \mathbf{H}_F$ and $\rm \mathbf{L}_F$. HF-MOE and LF-MOE share the similar structure. 
To be specific, the high-frequency feature extraction process for HF-MOE is defined as follows:
\begin{align}
&\rm \mathbf{W}_h = \mathbf{Gate}(\rm \mathbf{F}_h)\\
&\rm \mathbf{H}_F = \sum_{\rm i=1}^{\rm N} \rm \mathbf{W}_h^{i} \cdot \mathbf{\phi_{i}}({F_h})
\end{align}
In this context, the function $ \mathbf{Gate}(.)$ produces the gate weights $\rm \mathbf{W}_h \in \mathbb{R}^{\rm N}$, which are then utilized as the weighting coefficients for a linear combination of the different HF-EXPERT outputs. Here, $\mathbf{\phi_{i}(.)}$ represents the i-th HF-EXPERT block.
Depending on the input, the $\mathbf{Gate(.)}$ generates different weights, allowing the network structure to be dynamically adjusted. The details of $\mathbf{Gate(.)}$ will be discussed in the Experts Mixture part.
\begin{figure}
    \centering
    \includegraphics[width=\linewidth]{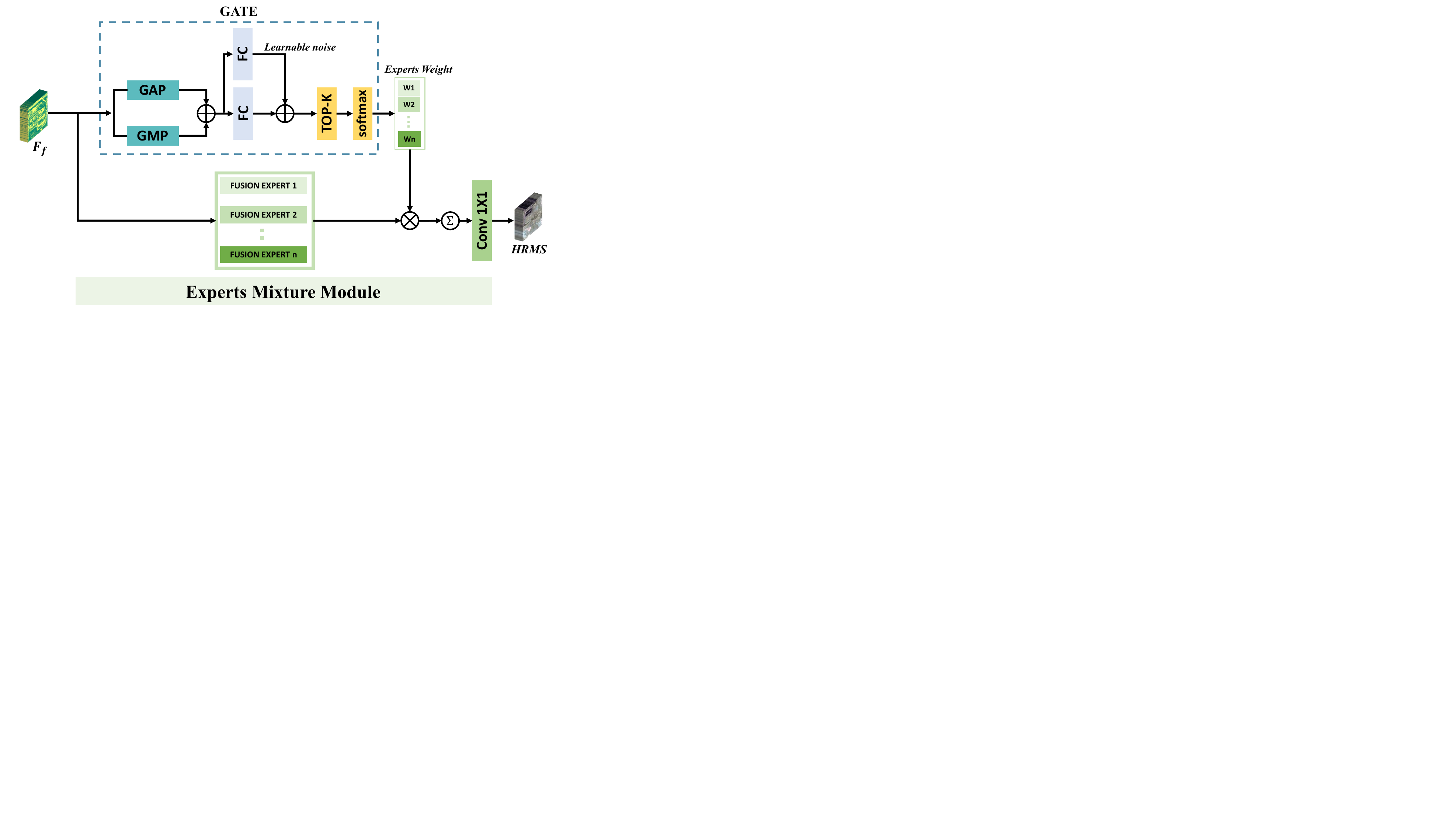}
    \caption{The architecture of Experts Mixture module, which includes the gating mechanism responsible for generating sparse weights based on input features, and the selection of appropriate fusion expert outputs based on the weights.}
    \label{fig:fusionsub}
\end{figure}

\noindent\textbf{Experts Mixture module.}
We have designed the Experts Mixture module, as shown in Figure~\ref{fig:fusionsub}, which adopts the MOE architecture and leverages multiple frequency learning experts to fuse input features and adapt to the diverse content of remote sensing images. Gate generates different weights for feature fusion, selecting the most suitable experts based on the input feature. To prevent homogenization among experts, the gate generates sparse weights. Given the input feature $\rm \mathbf{F}_f$, which is obtained by concatenating $\rm \mathbf{F}_{ms}$, $\rm \mathbf{F}_{pan}$, $\rm \mathbf{L}_F$, and $\rm \mathbf{H}_F$, the weight generation process is defined as follows:
\begin{align}
&\rm \mathbf{F}_{e} = \rm \mathbf{GAP}(\rm \mathbf{F}_f)+\mathbf{GMP}(\rm \mathbf{F}_f)\\
&\epsilon = \mathbf{SoftPlus}(\rm \mathbf{A}_1 \times \rm \mathbf{F}_{e})\\
&\rm \mathbf{V} = \rm \mathbf{A}_2 \times \rm \mathbf{F}_{e}+\epsilon\\
&\rm \mathbf{W}_f = \mathbf{Softmax} \circ\mathop{Topk}(\rm \mathbf{V})
\end{align}
Here, $\mathbf{GAP(.)}$ and $\mathbf{GMP(.)}$ correspond to average pooling and maximum pooling operations, respectively, while $\rm \mathbf{A}_1$ and $\rm \mathbf{A}_2$ are learnable matrices, specifically the fully connected layers shown in the Figure~\ref{fig:fusionsub}. First, we process the features using $\mathbf{GAP(.)}$ and $\mathbf{GMP(.)}$, and then we sum them to obtain $\rm \mathbf{F}_e$. Next, $\rm \mathbf{F}_e$ is passed through a fully connected layer, and learnable noise $\epsilon$ is added to produce $\rm \mathbf{V} \in \mathbb{R}^{N}$. By applying the Top-K operation, we select the k positions with the highest weight value from among the n weights, assign negative infinity to the remaining positions, and finally obtain the expert weight $\rm \mathbf{W}_f$ using $\mathbf{softmax(.)}$, with unselected expert weights reset to 0. The use of learnable noise ensures that the probability of each expert being selected can be made almost equal.

To obtain the output HRMS image, we first multiply the output of Expert weight and frequency learning experts, and then adjust the channel through convolution. The process can be defined as follows:
\begin{align}
&\rm \mathbf{HRMS} = \rm \mathbf{C}_1\sum_{\rm i=1}^{\rm N} \rm \mathbf{W}_f^{i} \cdot \mathbf{\psi_{i}}(\rm \mathbf{F}_f)
\end{align}
Here, $\mathbf{\psi_{i}}$ represents the i-th fusion expert, and $\rm \mathbf{C}_1$ is the convolution block with a $1 \times 1$ kernel size.
\subsection{Loss Function}
Our loss function comprises three parts: reconstruction loss, mask loss, and load loss. Reconstruction loss is used to minimize the difference between the model output and the ground truth. Mask loss is used to learn the appropriate frequency mask, and load loss balances the load of different experts in MOE and prevents some experts from being ignored during training.
Let the model output be denoted as $Y$, the ground truth as $G$, and the reconstruction loss as the L1 loss between the two, given by:
\begin{align}
\mathcal{L}_{rec} = ||\rm \mathbf{Y}-\mathbf{G}||_1
\end{align}
For the mask learning process, we first follow the procedure shown in Figure~\ref{fig:dctmoti} and use manually selected rough thresholds to generate the frequency mask label of the training data, which we refer to as $M_{gt}$. We minimize the L1 distance between the mask output from the mask predictor and $M_{gt}$. We adopt an annealing strategy to adjust the weight of the mask loss, ensuring that the network no longer relies on the mask label to generate more accurate masks after learning mask information. The mask loss is defined as follows:
\begin{align}
\mathcal{L}_{mask} = ||\rm \mathbf{M}-\mathbf{M}_{gt}||_1
\end{align}
To balance the load of experts, we use the square of the coefficient of variation (SCV) as the load loss. Given the weight $W$, SCV can be computed as:
\begin{align}
&\rm \mathbf{SCV}(W) = (\sigma(W)/\bar{W})^2
\end{align}
Here, $\sigma(W)$ and $\bar{W}$ denote the standard deviation and mean of the elements in the weight vector, respectively.

\noindent To balance the workload of the experts in our network, we use a load loss that is the sum of the SCV values of their respective weight vectors. This is given by:
\begin{align}
  \rm  \mathcal{L}_{load} = \rm \mathbf{SCV}(\rm \mathbf{W}_h) + \rm \mathbf{SCV}(\rm \mathbf{W}_l) + \rm \mathbf{SCV}(\rm \mathbf{W}_f)
\end{align}
where $\rm \mathbf{W}_h$, $\rm \mathbf{W}_l$, and $\rm \mathbf{W}_f$ represent the weight vectors of the HF-MOE, LF-MOE, and Experts Mixture module, respectively.

The total loss function is given by:
\begin{align}
\mathcal{L} = \rm \mathcal{L}_{rec}+ \alpha \mathcal{L}_{mask}+ \beta \mathcal{L}_{load}
\end{align}
where the initial value of $\alpha$ is 0.001, which is attenuated using an annealing strategy. After 70$\%$ of the training epochs, $\alpha$ decreases to 0 while $\beta$ is set to 0.1.

\begin{table*}[!ht]
	\centering
	\normalsize
%  	\renewcommand{\tabcolsep}{2.9pt} % adjust horizontal space
% \renewcommand{\arraystretch}{1.2}
% \resizebox{\linewidth}{!}{
\scalebox{0.86}{
\begin{tabular}{l|cllc|cllc|cllc}
\hline
                         & \multicolumn{4}{c|}{WorldView-II}                                                                                                                                                    & \multicolumn{4}{c|}{GaoFen2}                                                                                                                                                                  & \multicolumn{4}{c}{Worldview-III}                                                                                                                                                             \\ \cline{2-13} 
\multirow{2}{*}{Method} & \multicolumn{1}{l|}{PSNR$\uparrow$}                           & \multicolumn{1}{l|}{SSIM$\uparrow$}                          & \multicolumn{1}{l|}{SAM$\downarrow$}                           & ERGAS$\downarrow$                         & \multicolumn{1}{l|}{PSNR$\uparrow$}                           & \multicolumn{1}{l|}{SSIM$\uparrow$}                          & \multicolumn{1}{l|}{SAM$\downarrow$}                           & ERGAS$\downarrow$                         & \multicolumn{1}{l|}{PSNR$\uparrow$}                           & \multicolumn{1}{l|}{SSIM$\uparrow$}                          & \multicolumn{1}{l|}{SAM$\downarrow$}                           & ERGAS$\downarrow$                         \\ \hline
PANNET                   & \multicolumn{1}{l|}{40.8176}                        & \multicolumn{1}{l|}{0.9626}                        & \multicolumn{1}{l|}{0.0257}                        & 1.0557                        & \multicolumn{1}{l|}{43.0659}                        & \multicolumn{1}{l|}{0.9685}                        & \multicolumn{1}{l|}{0.0178}                        & 0.8577                        & \multicolumn{1}{l|}{29.6840}                        & \multicolumn{1}{l|}{0.9072}                        & \multicolumn{1}{l|}{0.0851}                        & 3.4263                        \\
MSDCNN                   & \multicolumn{1}{l|}{41.3355}                        & \multicolumn{1}{l|}{0.9664}                        & \multicolumn{1}{l|}{0.0242}                        & 0.9940                        & \multicolumn{1}{l|}{45.6847}                        & \multicolumn{1}{l|}{0.9827}                        & \multicolumn{1}{l|}{0.0135}                        & 0.6389                        & \multicolumn{1}{l|}{30.3038}                        & \multicolumn{1}{l|}{0.9184}                        & \multicolumn{1}{l|}{0.0782}                        & 3.1884                        \\
SRPPNN                   & \multicolumn{1}{l|}{41.4538}                        & \multicolumn{1}{l|}{0.9679}                        & \multicolumn{1}{l|}{0.0233}                        & 0.9899                        & \multicolumn{1}{l|}{47.1998}                        & \multicolumn{1}{l|}{0.9877}                        & \multicolumn{1}{l|}{0.0106}                        & 0.5586                        & \multicolumn{1}{l|}{30.4346}                        & \multicolumn{1}{l|}{0.9202}                        & \multicolumn{1}{l|}{0.0770}                        & 3.1553                        \\
GPPNN                    & \multicolumn{1}{l|}{41.1622}                        & \multicolumn{1}{l|}{0.9684}                        & \multicolumn{1}{l|}{0.0244}                        & 1.0315                        & \multicolumn{1}{l|}{44.2145}                        & \multicolumn{1}{l|}{0.9815}                        & \multicolumn{1}{l|}{0.0137}                        & 0.7361                        & \multicolumn{1}{l|}{30.1785}                        & \multicolumn{1}{l|}{0.9175}                        & \multicolumn{1}{l|}{0.0776}                        & 3.2593                        \\ 
MutNet                    & \multicolumn{1}{l|}{41.6773}                        & \multicolumn{1}{l|}{0.9705}                        & \multicolumn{1}{l|}{0.0224}                        & 0.9519                        & \multicolumn{1}{l|}{47.3042}                        & \multicolumn{1}{l|}{0.9892}                        & \multicolumn{1}{l|}{0.0102}                        & 0.5481                        & \multicolumn{1}{l|}{30.4907}                        & \multicolumn{1}{l|}{0.9223}                        & \multicolumn{1}{l|}{0.0749}                        & 3.1125                        \\ 
INNformer                      & \multicolumn{1}{l|}{41.6903}                        & \multicolumn{1}{l|}{0.9704}                        & \multicolumn{1}{l|}{0.0227}                        & 0.9514                      & \multicolumn{1}{l|}{47.3528}                        & \multicolumn{1}{l|}{0.9893}                        & \multicolumn{1}{l|}{0.0102}                        & 0.5479                        & \multicolumn{1}{l|}{30.5365}                        & \multicolumn{1}{l|}{0.9225}                        & \multicolumn{1}{l|}{0.0747}                        & 3.0997                        \\
SFINet                      & \multicolumn{1}{l|}{{41.7244}}                        & \multicolumn{1}{l|}{{\textbf{0.9725}}}                        & \multicolumn{1}{l|}{{0.0220}}                        &{0.9506}                      & \multicolumn{1}{l|}{{47.4712}}                        & \multicolumn{1}{l|}{{\textbf{0.9901}}}                        & \multicolumn{1}{l|}{{0.0102}}                        & {0.5479}                        & \multicolumn{1}{l|}{{30.5901}}                        & \multicolumn{1}{l|}{{0.9236}}                        & \multicolumn{1}{l|}{{0.0741}}                        & {3.0798}                        \\
\hline
Ours                     & \multicolumn{1}{l|}{\textbf{42.0262}} & \multicolumn{1}{l|}{{0.9723}} & \multicolumn{1}{l|}{\textbf{0.0215}} & {\textbf{0.9172}} & \multicolumn{1}{l|}{\textbf{47.6721}} & \multicolumn{1}{l|}{{0.9898}} & \multicolumn{1}{l|}{\textbf{0.0098}} & {\textbf{0.5242}} & \multicolumn{1}{l|}{\textbf{30.9903}} & \multicolumn{1}{l|}{\textbf{0.9287}} & \multicolumn{1}{l|}{\textbf{0.0697}} & {\textbf{2.9531}} \\ \hline
\end{tabular}}
% }
% }
		\caption{Quantitative comparison on three datasets. Best results highlighted in \textbf{bold}.
  % Best results are highlighted by \textcolor{red}{red} the second best resulare highlight by \textcolor{blue}{blue}. 
  $\uparrow$ indicates that the larger the value, the better the performance, and $\downarrow$ indicates that the smaller the value, the better the performance.}
\label{metric}
\end{table*}
\begin{table*}[!h]
\normalsize
\centering
% \renewcommand\arraystretch{1.2}
% \renewcommand{\tabcolsep}{3.2pt}

% The table highlights the best results in \textcolor{red}{red} color.}
% {
\scalebox{0.79}{
\begin{tabular}{c|ccccccccccccc}
\hline
Metric & Brovey &GS & IHS & GFPCA & PNN & PANNET & MSDCNN & SRPPNN & GPPNN &MutNet &INNformer &SFINet & Ours \\
\hline
$D_{\lambda}$ $\downarrow$ & 0.1378 & 0.0696 & 0.0770 & 0.0914 & 0.0746 &  0.0737 & 0.0734 & 0.0767 & 0.0782 & 0.0694 &0.0697&0.0681 &\textbf{0.0674} \\
$D_{S}$ $\downarrow$  & 0.2605 & 0.2456 & 0.2985 & 0.1635 & 0.1164 & 0.1224 & 0.1151 &  0.1162 & 0.1253 &\textbf{0.1118} &0.1128&0.1119 &0.1121 \\
% $D_s$\downarrow & 0.1164 & 0.1224 & 0.1151 & 0.1162 & 0.1253 & \textbf{0.1146} \\
QNR $\uparrow$ & 0.6390 & 0.0725 & 0.6485 & 0.7615 & 0.8191 & 0.8143 &  0.8215 & 0.8173 & 0.8073 &0.8259 &0.8253 &0.8276 &\textbf{0.8291} \\ \hline
\end{tabular}}
% }
\caption{Evaluation of the proposed method on real-world full-resolution scenes from the GaoFen2 dataset.}
\label{full}
\end{table*}
\section{Experiments}
\subsection{Datasets and Benchmark}
Our experiments were conducted on three typical datasets, namely WorldView-II (WV2), Gaofen2 (GF2), and WorldView-III (WV3), which include various natural and urban scenarios. Since ground truth is unavailable, we follow the Wald protocol~\cite{gt} to generate training samples. 
We compare our proposed method with cutting-edge deep learning methods, including PANNET, MSDCNN, SRPPNN, GPPNN, MutNet, INN-former and SFINet and some classic methods such as GFPCA~\cite{GFPCA}, GS~\cite{GS}, IHS~\cite{IHS}, Brovey~\cite{Brovey} and SFIM~\cite{SFIM}.
\subsection{Implement Details}
We trained our model using the Python framework on an RTX 3060 GPU, with four experts (n=4), k=2 for each MOE, and a total of 1000 epochs. We used the Adam optimizer with a learning rate of 5e-4 and linearly decreased the weight of the loss ($\alpha$) during training. For training samples, we cropped LRMS patches of size 32x32 and PAN patches of size 128x128 from the images, with a batch size of 4.
We have employed a comprehensive set of evaluation metrics to assess the performance of our approach, including well-established measures such as PSNR, SSIM, SAM~\cite{sam}, and ERGAS, as well as non-reference metrics such as $D_s$, $D_\lambda$ and QNR to evaluate the generalization performance of our model.
\subsection{Comparison with state-of-the-art methods}
\begin{figure*}
	\centering
	\includegraphics[width=\textwidth]{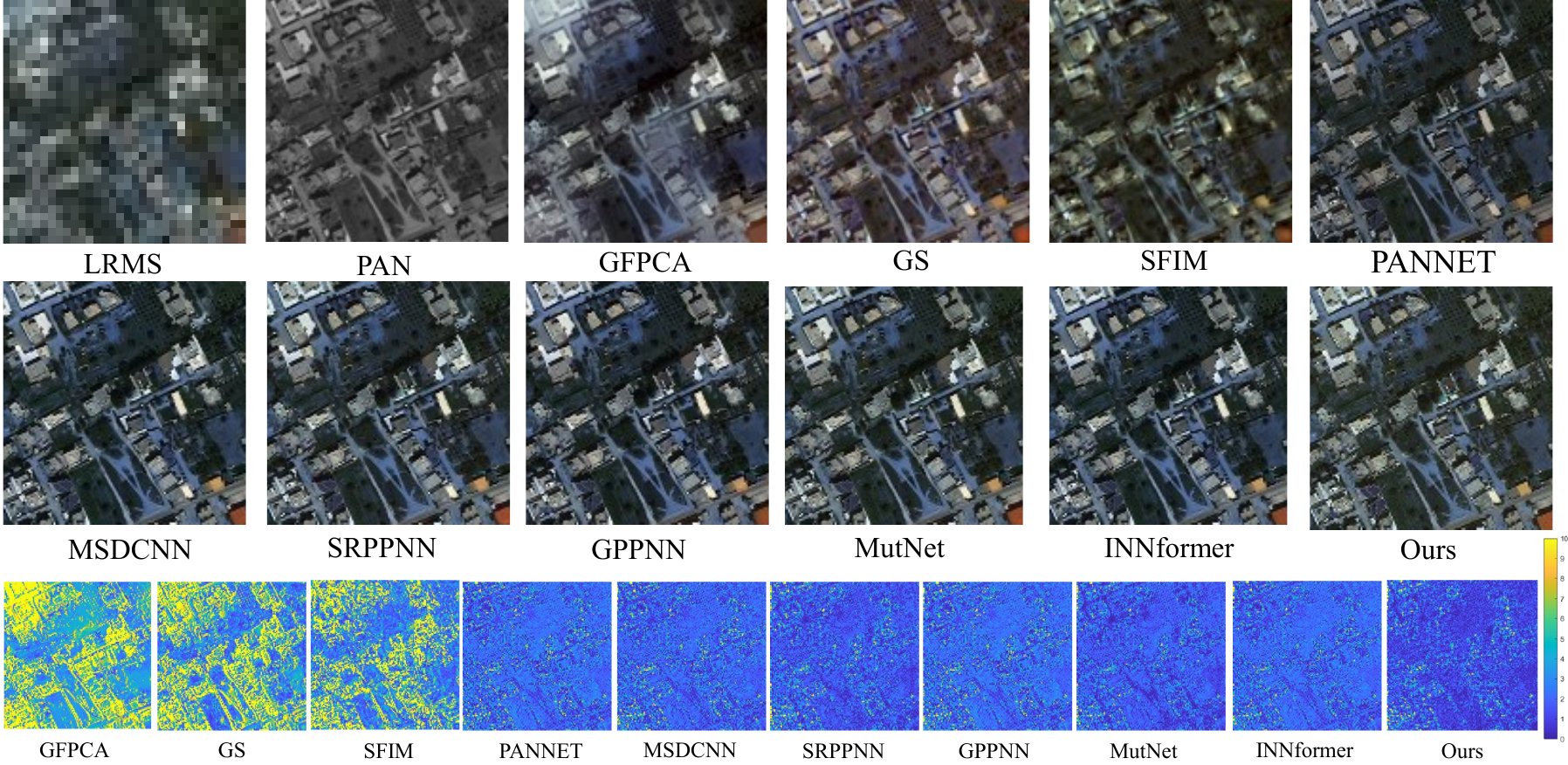}
	\caption{The result of our approach was compared against nine other methods on WorldView-III dataset.}
	\label{wv3}
\end{figure*}

\noindent\textbf{Evaluation on reduced-resolution scene.} The evaluation results on the three datasets are presented in Table~\ref{metric}, clearly demonstrating the superior performance of our proposed method over the SOTA methods in all metrics. Our model exhibits significant improvements in PSNR across all three datasets, with 0.301dB improvement on WV2 and 0.4dB improvement on WV3 compared to the INNformer, respectively. These results validate the consistency of our method with the ground truth, and other metrics further confirm the effectiveness of our approach. Qualitative experiments in Figure~\ref{wv3} showcase representative samples from the WV3 datasets, where the residual plot of our method has the least brightness, indicating its closeness to the ground truth. Our method provides clear edges and accurate spectra, further highlighting its superiority over other methods.

\begin{figure*}[ht!]
	\centering
	\includegraphics[width=\textwidth]{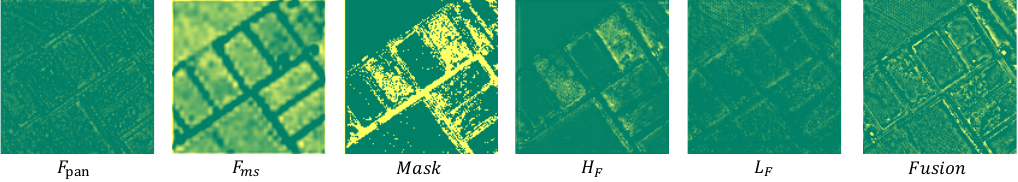}
	\caption{The network's feature map.}
	\label{fmi}
\end{figure*}

\begin{table*}[!h]
	\normalsize
\centering
% \resizebox{\linewidth}{!}{
\scalebox{0.72}{
\begin{tabular}{ccc|cllc|cllc|cllc}
\hline
                         &                               &                       & \multicolumn{4}{c|}{WorldView-II}                                                                                              & \multicolumn{4}{c|}{GaoFen2}                                                                                                   & \multicolumn{4}{c}{WorldView-III}                                                                                              \\ \cline{4-15} 
\multirow{-2}{*}{Config} & \multirow{-2}{*}{Mask} & \multirow{-2}{*}{Experts Mixture} & PSNR$\uparrow$                           & SSIM$\uparrow$                          & SAM$\downarrow$                          & ERGAS$\downarrow$                          & PSNR$\uparrow$                           & SSIM$\uparrow$                          & SAM$\downarrow$                            & ERGAS$\downarrow$                          & PSNR$\uparrow$                           & SSIM$\uparrow$                          & SAM$\downarrow$                            & ERGAS$\downarrow$                          \\ \hline
(I)                      & \XSolid                        & \Checkmark             & 41.8998                        & 0.9720                        & 0.0220                        & 0.9322                        & 47.5914                        & 0.9896                        & 0.0100                        & 0.5312                        & 30.8387                        & 0.9261                        & 0.0720 & 2.9969                        \\
(II)                     & \Checkmark                     & \XSolid                & 41.8274                        & 0.9714                       & 0.0220                        & 0.9358                        & 47.3596                        & 0.9888                        & 0.0102                        & 0.5429                        & 30.7930                        & 0.9261                        & 0.0717                        & 3.0134                        \\ \cline{1-3}
Ours                     & \Checkmark                     & \Checkmark             & {\textbf{ 42.0261}} & {\textbf{0.9723}} & {\textbf{0.0215}} & {\textbf{ 0.9172}} & {\textbf{47.6721}} & {\textbf{0.9898}} & {\textbf{ 0.0098}} & {\textbf{ 0.5242}} & {\textbf{30.9903}} & {\textbf{0.9287}} & {\textbf{0.0697}}                        & {\textbf{ 2.9531}} \\ \hline
\end{tabular}}
% }
\caption{The results of the ablation experiments conducted on the three datasets.}\label{abl}

\end{table*}
\noindent\textbf{Evaluation on Full-Resolution Scene.} To assess the generalization ability of our method, we evaluated it on the full Gaofen2 dataset using no-reference metrics. This dataset consists of images from the reserved part of the Gaofen2 dataset, which were not downsampled. The experimental results, as shown in Table~\ref{full}, demonstrate that our method outperformed other approaches on all three metrics, indicating the exceptional adaptability of the MOE architecture to remote sensing images.
\subsection{Ablation Experiments}
The core components of our network comprise of the Mask Predictor, Frequency Experts module, and Experts Mixture module. The former two are responsible for improving the network's frequency domain perception, while the latter enables dynamic feature fusion. We conducted two sets of ablation experiments on three datasets, the results of which are presented in Table~\ref{abl}. For more ablation experiments, please refer to the supplementary material.

\noindent\textbf{Mask Predictor.} The mask predictor serves as the core component for frequency domain perception. In the first set of experiments, we conducted ablation by removing the mask predictor and the split operation, and directly feed the $F_c$ into LF-MOE and HF-MOE. Due to the loss of the frequency mask, both of them were unable to process high and low-frequency information in a targeted manner. The experimental results, shown in the first row of Table~\ref{abl}, demonstrate a significant decrease in various indicators, proving that the targeted processing of high-frequency and low-frequency information can promote the network's learning of high-frequency information to improve detail perception.

\noindent\textbf{Experts Mixture module.} In the second set of experiments, we replaced the Experts Mixture module with a resblock having similar parameter number for feature fusion, thereby eliminating the dynamic feature fusion capability of the network. The experimental results in the second row of the Table~\ref{abl} clearly demonstrate that the deletion of the Experts Mixture module led to a decrease in various evaluation indicators on the three datasets. This indicates the significant role played by dynamic network structures in processing diverse remote sensing images.
\subsection{Visualization of Feature Maps}
To further illustrate the capabilities of our model, we have visualized the feature maps generated by our network, as presented in the Figure~\ref{fmi}. The columns from left to right show the PAN image feature maps $\rm \mathbf{F}_{pan}$, MS image feature maps $\rm \mathbf{F}_{ms}$, Mask predictor output $\rm \mathbf{M}$, HF-MOE output $\rm \mathbf{H}_F$, LF-MOE output $\rm \mathbf{L}_F$, and Experts Mixture module output before channel adjustment, respectively. By observing the feature maps, it is evident that the mask predicted by the network can accurately distinguish between the high and low frequency components of remote sensing images, which is much more precise than the masks generated by the artificial threshold selection method. Moreover, it is clear from the $\rm \mathbf{H}_F$ and $\rm \mathbf{L}_F$ feature maps that the HF-MOE and LF-MOE components specifically learn the high and low frequency information of the image, respectively. Finally, the Experts Mixture module effectively integrates all the information. These feature maps demonstrate the targeted processing of information at different frequencies by our network.
\section{Conclusion}
This work introduces a new approach that employs a frequency mask for managing high and low-frequency data and a dynamic structure to adjust to the various content of remote sensing images. The proposed method utilizes a Frequency Adaptive Mixture of Experts (MOE) network, which is designed to target both high and low-frequency data while employing a dynamic network structure. Notably, our research is the first to apply the MOE structure in pan-sharpening. Extensive experiments indicate that our model outperforms state-of-the-art methods and exhibits robust generalization capabilities.
\section{Acknowledgement}
This work was Supported by the Natural Science Foundation of Anhui Province (No.2208085MC57), and HFIPS Director’s Fund, Grant No.2023YZGH04.
\bibliography{aaai24}

\begin{thebibliography}{36}
\providecommand{\natexlab}[1]{#1}

\bibitem[{Ahmed, Natarajan, and Rao(1974)}]{ahmed1974discrete}
Ahmed, N.; Natarajan, T.; and Rao, K.~R. 1974.
\newblock Discrete cosine transform.
\newblock \emph{IEEE transactions on Computers}, 100(1): 90--93.

\bibitem[{Cai and Huang(2021)}]{srppnn}
Cai, J.; and Huang, B. 2021.
\newblock Super-Resolution-Guided Progressive Pansharpening Based on a Deep Convolutional Neural Network.
\newblock \emph{IEEE Transactions on Geoscience and Remote Sensing}, 59(6): 5206--5220.

\bibitem[{Cao et~al.(2023)Cao, Sun, Zhu, and Hu}]{cao2023multi}
Cao, B.; Sun, Y.; Zhu, P.; and Hu, Q. 2023.
\newblock Multi-Modal Gated Mixture of Local-to-Global Experts for Dynamic Image Fusion.
\newblock In \emph{Proceedings of the IEEE/CVF International Conference on Computer Vision}, 23555--23564.

\bibitem[{Chen et~al.(2021)Chen, Lu, Zhang, Chu, and Chen}]{Chen_2021_CVPR}
Chen, L.; Lu, X.; Zhang, J.; Chu, X.; and Chen, C. 2021.
\newblock HINet: Half Instance Normalization Network for Image Restoration.
\newblock In \emph{Proceedings of the IEEE/CVF Conference on Computer Vision and Pattern Recognition (CVPR) Workshops}, 182--192.

\bibitem[{Dai et~al.(2021)Dai, Li, Liu, Tong, and Duan}]{dai2021generalizable}
Dai, Y.; Li, X.; Liu, J.; Tong, Z.; and Duan, L.-Y. 2021.
\newblock Generalizable person re-identification with relevance-aware mixture of experts.
\newblock In \emph{Proceedings of the IEEE/CVF Conference on Computer Vision and Pattern Recognition}, 16145--16154.

\bibitem[{Dong et~al.(2016)Dong, Loy, He, and Tang}]{srcnn}
Dong, C.; Loy, C.~C.; He, K.; and Tang, X. 2016.
\newblock Image Super-Resolution Using Deep Convolutional Networks.
\newblock \emph{IEEE Transactions on Pattern Analysis and Machine Intelligence}, 38(2): 295--307.

\bibitem[{Fasbender, Radoux, and Bogaert(2008)}]{fasbender2008bayesian}
Fasbender, D.; Radoux, J.; and Bogaert, P. 2008.
\newblock Bayesian data fusion for adaptable image pansharpening.
\newblock \emph{IEEE Transactions on Geoscience and Remote Sensing}, 46(6): 1847--1857.

\bibitem[{Fuoli, Van~Gool, and Timofte(2021)}]{fuoli2021fourier}
Fuoli, D.; Van~Gool, L.; and Timofte, R. 2021.
\newblock Fourier space losses for efficient perceptual image super-resolution.
\newblock In \emph{Proceedings of the IEEE/CVF International Conference on Computer Vision}, 2360--2369.

\bibitem[{Gillespie, Kahle, and Walker(1987)}]{Brovey}
Gillespie, A.~R.; Kahle, A.~B.; and Walker, R.~E. 1987.
\newblock Color enhancement of highly correlated images. II. Channel ratio and "chromaticity" transformation techniques - ScienceDirect.
\newblock \emph{Remote Sensing of Environment}, 22(3): 343--365.

\bibitem[{Gross, Ranzato, and Szlam(2017)}]{gross2017hard}
Gross, S.; Ranzato, M.; and Szlam, A. 2017.
\newblock Hard mixtures of experts for large scale weakly supervised vision.
\newblock In \emph{Proceedings of the IEEE Conference on Computer Vision and Pattern Recognition}, 6865--6873.

\bibitem[{Haydn et~al.(1982)Haydn, Dalke, Henkel, and Bare}]{IHS}
Haydn, R.; Dalke, G.~W.; Henkel, J.; and Bare, J.~E. 1982.
\newblock Application of the IHS color transform to the processing of multisensor data and image enhancement.
\newblock \emph{National Academy of Sciences of the United States of America}, 79(13): 571--577.

\bibitem[{Jang, Gu, and Poole(2017)}]{jang2017categorical}
Jang, E.; Gu, S.; and Poole, B. 2017.
\newblock Categorical Reparameterization with Gumbel-Softmax.
\newblock In \emph{International Conference on Learning Representations}.

\bibitem[{Jordan and Jacobs(1994)}]{jordan1994hierarchical}
Jordan, M.~I.; and Jacobs, R.~A. 1994.
\newblock Hierarchical mixtures of experts and the EM algorithm.
\newblock \emph{Neural computation}, 6(2): 181--214.

\bibitem[{Laben and Brower(2000)}]{GS}
Laben, C.; and Brower, B. 2000.
\newblock Process for Enhancing the Spatial Resolution of Multispectral Imagery Using Pan-Sharpening.
\newblock \emph{US Patent 6011875A}.

\bibitem[{Liao et~al.(2017)Liao, Xin, Coillie, Thoonen, and Philips}]{GFPCA}
Liao, W.; Xin, H.; Coillie, F.~V.; Thoonen, G.; and Philips, W. 2017.
\newblock Two-stage fusion of thermal hyperspectral and visible RGB image by PCA and guided filter.
\newblock In \emph{Workshop on Hyperspectral Image and Signal Processing: Evolution in Remote Sensing}.

\bibitem[{Liu.(2000)}]{SFIM}
Liu., J.~G. 2000.
\newblock Smoothing filter-based intensity modulation: A spectral preserve image fusion technique for improving spatial details.
\newblock \emph{International Journal of Remote Sensing}, 21(18): 3461--3472.

\bibitem[{Magid et~al.(2021)Magid, Zhang, Wei, Jang, Lin, Fu, and Pfister}]{magid2021dynamic}
Magid, S.~A.; Zhang, Y.; Wei, D.; Jang, W.-D.; Lin, Z.; Fu, Y.; and Pfister, H. 2021.
\newblock Dynamic high-pass filtering and multi-spectral attention for image super-resolution.
\newblock In \emph{Proceedings of the IEEE/CVF International Conference on Computer Vision}, 4288--4297.

\bibitem[{Mallat(1989)}]{DWT1989}
Mallat, S. 1989.
\newblock A Theory for Multiresolution Signal Decomposition: The Wavelet Representation.
\newblock \emph{IEEE Transactions on Pattern Analysis and Machine Intelligence}, 11(7): 674--693.

\bibitem[{Masi et~al.(2016)Masi, Cozzolino, Verdoliva, and Scarpa}]{pnn}
Masi, G.; Cozzolino, D.; Verdoliva, L.; and Scarpa, G. 2016.
\newblock Pansharpening by convolutional neural networks.
\newblock \emph{Remote Sensing}, 8(7): 594.

\bibitem[{Nunez et~al.(1999)Nunez, Otazu, Fors, Prades, Pala, and Arbiol}]{ATWT1999}
Nunez, J.; Otazu, X.; Fors, O.; Prades, A.; Pala, V.; and Arbiol, R. 1999.
\newblock Multiresolution-based image fusion with additive wavelet decomposition.
\newblock \emph{IEEE Transactions on Geoscience and Remote sensing}, 37(3): 1204--1211.

\bibitem[{Palsson, Sveinsson, and Ulfarsson(2013)}]{tv}
Palsson, F.; Sveinsson, J.~R.; and Ulfarsson, M.~O. 2013.
\newblock A new pansharpening algorithm based on total variation.
\newblock \emph{IEEE Geoscience and Remote Sensing Letters}, 11(1): 318--322.

\bibitem[{Schowengerdt(1980)}]{HPF}
Schowengerdt, R.~A. 1980.
\newblock Reconstruction of multispatial, multispectral image data using spatial frequency content.
\newblock \emph{Photogrammetric Engineering and Remote Sensing}, 46(10): 1325--1334.

\bibitem[{Shazeer et~al.(2017)Shazeer, Mirhoseini, Maziarz, Davis, Le, Hinton, and Dean}]{shazeer2017}
Shazeer, N.; Mirhoseini, A.; Maziarz, K.; Davis, A.; Le, Q.; Hinton, G.; and Dean, J. 2017.
\newblock Outrageously Large Neural Networks: The Sparsely-Gated Mixture-of-Experts Layer.
\newblock In \emph{International Conference on Learning Representations}.

\bibitem[{Vivone et~al.(2014)Vivone, Alparone, Chanussot, Dalla~Mura, Garzelli, Licciardi, Restaino, and Wald}]{LPTl}
Vivone, G.; Alparone, L.; Chanussot, J.; Dalla~Mura, M.; Garzelli, A.; Licciardi, G.~A.; Restaino, R.; and Wald, L. 2014.
\newblock A critical comparison among pansharpening algorithms.
\newblock \emph{IEEE Transactions on Geoscience and Remote Sensing}, 53(5): 2565--2586.

\bibitem[{Wald, Ranchin, and Mangolini(1997)}]{gt}
Wald, L.; Ranchin, T.; and Mangolini, M. 1997.
\newblock Fusion of satellite images of different spatial resolutions: Assessing the quality of resulting images.
\newblock \emph{Photogrammetric Engineering and Remote Sensing}, 63: 691--699.

\bibitem[{Xie et~al.(2021)Xie, Song, Xu, Xu, Zhang, and Wang}]{xie2021learning}
Xie, W.; Song, D.; Xu, C.; Xu, C.; Zhang, H.; and Wang, Y. 2021.
\newblock Learning frequency-aware dynamic network for efficient super-resolution.
\newblock In \emph{Proceedings of the IEEE/CVF International Conference on Computer Vision}, 4308--4317.

\bibitem[{Xu et~al.(2021)Xu, Zhang, Zhao, Sun, Liu, and Zhang}]{gppnn}
Xu, S.; Zhang, J.; Zhao, Z.; Sun, K.; Liu, J.; and Zhang, C. 2021.
\newblock Deep Gradient Projection Networks for Pan-sharpening.
\newblock In \emph{IEEE Conference on Computer Vision and Pattern Recognition}, 1366--1375.

\bibitem[{Yan et~al.(2022{\natexlab{a}})Yan, Zhou, Huang, Zhao, Xie, Li, and Hong}]{yan2022panchromatic}
Yan, K.; Zhou, M.; Huang, J.; Zhao, F.; Xie, C.; Li, C.; and Hong, D. 2022{\natexlab{a}}.
\newblock Panchromatic and Multispectral Image Fusion via Alternating Reverse Filtering Network.
\newblock \emph{Advances in Neural Information Processing Systems}, 35: 21988--22002.

\bibitem[{Yan et~al.(2022{\natexlab{b}})Yan, Zhou, Zhang, and Xie}]{yan2022memory}
Yan, K.; Zhou, M.; Zhang, L.; and Xie, C. 2022{\natexlab{b}}.
\newblock Memory-Augmented Model-Driven Network for Pansharpening.
\newblock In \emph{European Conference on Computer Vision}, 306--322. Springer.

\bibitem[{Yang et~al.(2017)Yang, Fu, Hu, Huang, Ding, and Paisley}]{yang2017pannet}
Yang, J.; Fu, X.; Hu, Y.; Huang, Y.; Ding, X.; and Paisley, J. 2017.
\newblock PanNet: A deep network architecture for pan-sharpening.
\newblock In \emph{IEEE International Conference on Computer Vision}, 5449--5457.

\bibitem[{Yuan et~al.(2018)Yuan, Wei, Meng, Shen, and Zhang}]{msdcnn}
Yuan, Q.; Wei, Y.; Meng, X.; Shen, H.; and Zhang, L. 2018.
\newblock A Multiscale and Multidepth Convolutional Neural Network for Remote Sensing Imagery Pan-Sharpening.
\newblock \emph{IEEE Journal of Selected Topics in Applied Earth Observations and Remote Sensing}, 11(3): 978--989.

\bibitem[{Yuhas, Goetz, and Boardman(1992)}]{sam}
Yuhas, R.~H.; Goetz, A.~F.; and Boardman, J.~W. 1992.
\newblock Discrimination among semi-arid landscape endmembers using the spectral angle mapper (SAM) algorithm.
\newblock In \emph{JPL, Summaries of the Third Annual JPL Airborne Geoscience Workshop. Volume 1: AVIRIS Workshop}.

\bibitem[{Zhang et~al.(2019)Zhang, Huang, Liu, and Tao}]{zhang2019learning}
Zhang, L.; Huang, S.; Liu, W.; and Tao, D. 2019.
\newblock Learning a mixture of granularity-specific experts for fine-grained categorization.
\newblock In \emph{Proceedings of the IEEE/CVF International Conference on Computer Vision}, 8331--8340.

\bibitem[{Zhou et~al.(2022{\natexlab{a}})Zhou, Huang, Fang, Fu, and Liu}]{zhou2022pan}
Zhou, M.; Huang, J.; Fang, Y.; Fu, X.; and Liu, A. 2022{\natexlab{a}}.
\newblock Pan-sharpening with customized transformer and invertible neural network.
\newblock In \emph{Proceedings of the AAAI Conference on Artificial Intelligence}, volume~36, 3553--3561.

\bibitem[{Zhou et~al.(2022{\natexlab{b}})Zhou, Huang, Yan, Yu, Fu, Liu, Wei, and Zhao}]{zhou2022spatial}
Zhou, M.; Huang, J.; Yan, K.; Yu, H.; Fu, X.; Liu, A.; Wei, X.; and Zhao, F. 2022{\natexlab{b}}.
\newblock Spatial-frequency domain information integration for pan-sharpening.
\newblock In \emph{European Conference on Computer Vision}, 274--291. Springer.

\bibitem[{Zhou et~al.(2022{\natexlab{c}})Zhou, Yan, Huang, Yang, Fu, and Zhao}]{zhou2022mutual}
Zhou, M.; Yan, K.; Huang, J.; Yang, Z.; Fu, X.; and Zhao, F. 2022{\natexlab{c}}.
\newblock Mutual information-driven pan-sharpening.
\newblock In \emph{Proceedings of the IEEE/CVF Conference on Computer Vision and Pattern Recognition}, 1798--1808.

\end{thebibliography}

\end{document}